\documentclass[conference]{IEEEtran}
\IEEEoverridecommandlockouts
\usepackage{cite}
\usepackage{amsmath,amssymb,amsfonts}
\usepackage{algorithmic}
\usepackage{graphicx}
\usepackage{textcomp}
\usepackage{tabularx,booktabs}
\usepackage{verbatim}
\usepackage[section]{placeins}

\usepackage{xcolor}
\def\BibTeX{{\rm B\kern-.05em{\sc i\kern-.025em b}\kern-.08em
    T\kern-.1667em\lower.7ex\hbox{E}\kern-.125emX}}
    
    \makeatletter
\newcommand{\linebreakand}{%
  \end{@IEEEauthorhalign}
  \hfill\mbox{}\par
  \mbox{}\hfill\begin{@IEEEauthorhalign}
}
\makeatother
\begin{document}

\title{Bengali Document Layout Analysis - A YOLOV8 Based Ensembling Approach
}

\author{
    \IEEEauthorblockN{Nazmus Sakib Ahmed\textsuperscript{*},
    Saad Sakib Noor\textsuperscript{*},
    Ashraful Islam Shanto Sikder\textsuperscript{*},
    Abhijit Paul}
    \IEEEauthorblockA{\textit{Inst. of Information Technology} \\
    \textit{University of Dhaka}\\
    Dhaka, Bangladesh \\
    Email: \{bsse1108, bsse1122, bsse1124, bsse1201\}@iit.du.ac.bd}
    \thanks{\textsuperscript{*}First three authors contributed equally to this research.}
}

\maketitle
\thispagestyle{plain}
\pagestyle{plain}

\begin{abstract}

This paper focuses on enhancing Bengali Document Layout Analysis (DLA) using the YOLOv8 model and innovative post-processing techniques. We tackle challenges unique to the complex Bengali script by employing data augmentation for model robustness. After meticulous validation set evaluation, we fine-tune our approach on the complete dataset, leading to a two-stage prediction strategy for accurate element segmentation.
Our ensemble model, combined with post-processing, outperforms individual base architectures, addressing issues identified in the BaDLAD dataset. By leveraging this approach, we aim to advance Bengali document analysis, contributing to improved OCR and document comprehension and BaDLAD serves as a foundational resource for this endeavor, aiding future research in the field. Furthermore, our experiments provided key insights to incorporate new strategies into the established solution.
\end{abstract}


\section{Introduction}
Document layout analysis (DLA) is a crucial step in many document processing tasks, such as optical character recognition (OCR) and document understanding \cite{dlm_survey_paper}. DLA involves the identification and classification of document elements, such as text boxes, paragraphs, images, and tables.

Bengali is a complex language with a rich and diverse script. This makes Bengali DLA a challenging task. \cite{complexity} In recent years, there has been significant progress in deep learning-based DLA for English. However, progress in Bangla on DLA is far behind.

The BaDLAD\cite{badlad} dataset is the first large-scale Bengali DLA dataset. It contains 33,695 human annotated document samples from six domains, with 710K polygon annotations for four unit types: text-box, paragraph, image, and table. 

In this paper, we used Yolo(You Only Look Once)\cite{yolo} model along with some post-processing for Bangla Document Object Modeling. We used a variety of data augmentation techniques to improve the robustness of our YOLOv8m-seg model. We evaluated our models on a validation set and selected the best performing model and configuration. Then we reproduced that model by training it on the whole dataset. We conducted manual testing on select test set images to identify and address issues with image and table masks. We used post-processing techniques to improve the quality of our predictions, such as convex hull filling \cite{convex_hull}. We used a two-stage prediction strategy to segment images and tables more accurately. The first stage used the general model to detect and segment all of the given instance classes. The second stage used the image model to specifically detect and segment images. We finally show that our ensemble approach with post-processing outperforms each base architecture by a significant margin.

We performed numerous experiments to address some of the issues mentioned in the BaDLAD paper along with some of the issues we observed while testing the models. Some of the strategies are incorporated into the solution presented in this paper.

We believe our work will help improve Bengali documents analysis, laying the pathway to more accurate OCR in the Bengali language.

% \section{Experiments}

\section{Methodology}
We go into detail about our approach and the construction of our solution in this section.

\begin{figure*}[h]
    \centering
    \setlength{\fboxrule}{0pt}
    \fbox{\includegraphics[width=\textwidth]{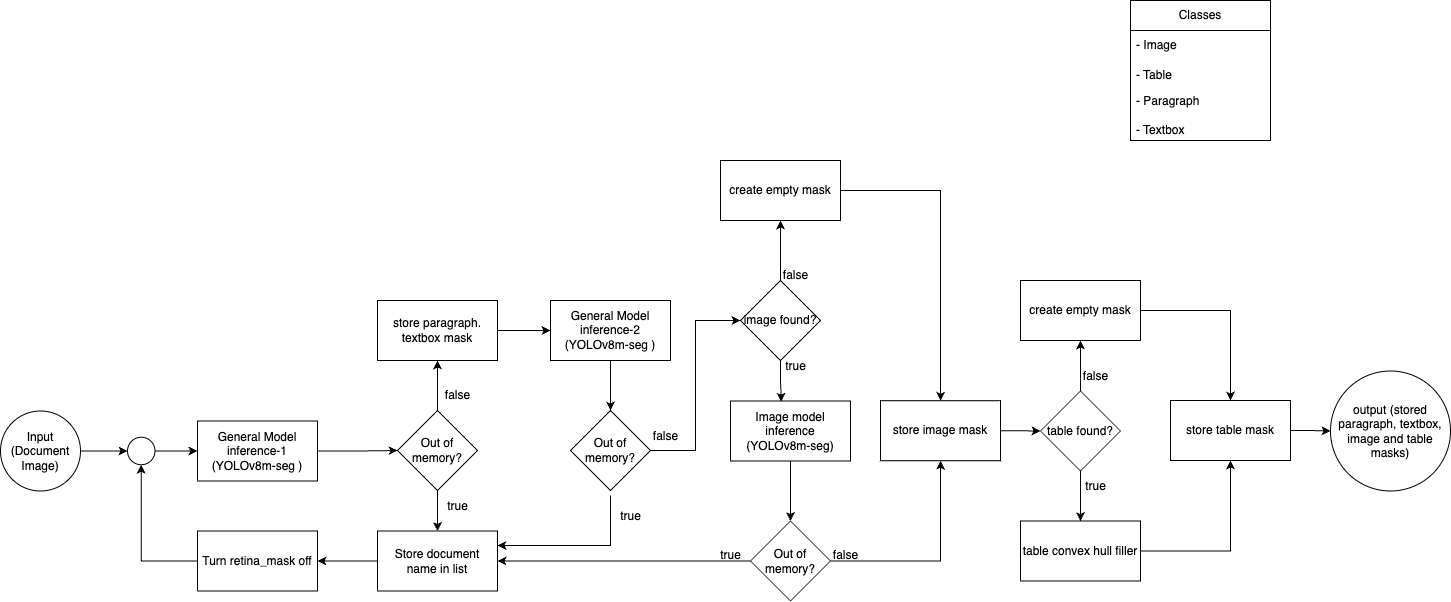}}
    \caption{High level view of inference}
    \label{fig: ensemble_structure}
\end{figure*}

\subsection{Dataset}
We utilized the BaDLAD \cite{badlad} - A Large Multi-Domain Bengali Document Layout Analysis Dataset, a comprehensive multi-domain collection of Bangla documents. BaDLAD contains various document types, including paragraphs, textboxes, images, and tables, making it suitable for training and evaluating our segmentation model across different document elements.

\subsection{Preprocessing and Augmentation}
\textbf{Training Data Augmentation:}
To enhance the robustness of our segmentation model, we applied specific data augmentation techniques during the training phase. We user YOLO's built-in augmentation interface to facilitate Mixup\cite{mixup}, copy-paste\cite{copypaste} and mosaic\cite{mosaic}, perspective\cite{perspective} augmentation on the training data.
Mixup, copy-paste, mosaic, and perspective augmentations are transformative data strategies in deep learning. Mixup blends images, copy-paste enriches with diverse segments, mosaic amalgamates scenes, and perspective adds spatial distortion. Employed in computer vision, these methods enhance model robustness, generalization, and performance.
% \textbf{Histogram Equalization:} We employed histogram equalization, which enhances the contrast of images by redistributing pixel values. This technique, ranked third in performance, contributes to better feature extraction and segmentation.

\textbf{Validation Set Construction:}
To facilitate model evaluation and selection, we created a validation set by separating 20\% of images for each class instance (totaling 4180 images) into a dedicated validation set. We ensured that this set contained appropriate representations for all classes. This split allowed us to monitor the performance of various models during training and identify the top performing models.

\textbf{Single Class Dataset:}
We selected all the documents containing image instances and removed all annotations except for image annotations. This augmented dataset also included 20\% of the documents without image instances from the original dataset. This augmented dataset was exclusively used for our image model, which was specifically designed for segmenting image instances. 

\subsection{Model Architecture and Configuration:}
We selected YOLOv8 as the model in our solution. \cite{yolo} It is one of the state-of-the-art vision model. It discarded the anchor-free model unlike its predecessors and this makes it suitable for instance segmentation tasks. Furthermore, this model produced much better baseline on the BADLAD-dataset\cite{badlad} compared to other alternatives. We selected the 'yolov8m-seg' variant of this model.
\subsection{Hyperparameters:}
After thorough testing and experimentation, the following hyperparameters were determined to yield optimal results for our selected model:
   \textbf{overlap\_mask:} True,
   \textbf{mask\_ratio:} 2,
   \textbf{augment:} True,
   \textbf{imgsz:} 672,
   \textbf{epochs:} 50,
   \textbf{copy\_paste:} 0.1,
   \textbf{mixup:} 0.1,
   \textbf{mosaic:} 0.5,
   \textbf{perspective:} 0.001

% \includegraphics[width=0.5\textwidth, height=6cm]{image-16.png}
% \label{YOLOv8 Model Architecture}

\subsection{Training:}
Our approach utilizes two models, as can be seen in Fig-\ref{fig: ensemble_structure}. A model to segment only images and another to detect the other classes. We used the validation set to train models of various configurations and to find optimal inference arguments. Only after sufficient testing we used the whole dataset to train our model for final inference. We only mention the final training instances in this section.

\subsubsection{\textbf{General model}} We finetuned our model on the whole dataset to detect and segment all of the given instance classes. We fine-tuned this model for 50 epochs. Our experiments showed that providing more training epochs than 50 degrades results.

\subsubsection{\textbf{Single class model for images}} A specialized YOLOv8m-seg model was trained on our dataset with the singular purpose of accurately recognizing and segmenting images. This deliberate focus on image segmentation emerged as a strategic response to address a critical shortcoming encountered when utilizing the general model for this task.

\vspace{10pt}

The image masks generated by the general model exhibited occasional deficiencies, most notably manifesting as gaps within the mask. This phenomenon became particularly apparent when images contained additional classes coexisting within the image mask. These gaps compromised the integrity and precision of the segmentation process, impeding our goal of obtaining meticulous image masks. 
We trained this model for 100 epochs on the single class image dataset.

\subsection{Testing to find optimal inference parameters:}
After the completion of model training, an extensive testing phase was conducted to comprehensively assess the strengths and weaknesses of each individual model and to compare their performance against other models.

\subsubsection{\textbf{Validation Metric Analysis}}
Models underwent rigorous evaluation on the validation set, utilizing diverse metrics including class-wise DICE\cite{dice} scores, instances prediction, correct predictions, predicted area, intersection area, and mAP50 \cite{map50_score}. Metrics were computed across 20 image size and confidence threshold combinations for each of the 10 models, yielding over 200 scores. 

These scores revealed optimal image size and confidence thresholds for each class. Notably, image sizes of 640px generally excelled, with different confidence thresholds impacting distinct classes. Lower confidences (e.g., 0.25) enhanced paragraph and textbox performance, while higher confidences (e.g., 0.45) benefited tables. Images performed best at around 0.35 confidence. This knowledge prompted separate inferences for each class with its optimal threshold. Validation scores also illuminated post-processing effects.

Integrating validation and submission scores pinpointed overfitting onset, guiding dataset-wide model replication to address overfitting concerns.

\subsubsection{\textbf{Manual Testing}}
In addition to the quantitative evaluation, a qualitative manual testing procedure was executed on select test set images. This manual scrutiny provided a nuanced understanding of the models' strengths and weaknesses. Particularly, it unveiled issues within image and table masks that emerged when these entities intersected with other classes. The resulting gaps or inaccuracies in image and table masks negatively impacted their respective scores. This observation prompted the formulation of strategies to address such challenges.

\subsection{Inference}
\subsubsection{\textbf{Prediction}}
Two distinct prediction strategies were employed on the general model, each tailored to the characteristics of specific document elements. And for the we used the image model on top of the general model to enhance image prediction

\textbf{Paragraph \& Textbox Prediction}

For the identification of paragraphs and textboxes, predictions were made with a confidence threshold of 0.25. This threshold ensured a balanced approach, effectively capturing instances of these entities within documents.

\textbf{Image \& Table Prediction}

Predictions for image and table classes were conducted with a higher confidence threshold of 0.35. This elevated threshold optimized the model's ability to precisely discern image and table elements within the document.

\textbf{Key Parameters}

Several critical parameters guided the predictive process:

\begin{itemize}
  \item \textbf{retina\_masks:} Enabled retina masks \cite{retina_mask} for enhanced segmentation precision. Enabling this feature makes the model use high resolution segmentation masks. 
  \item \textbf{conf:} Confidence thresholds in the general model were set at 0.25 for paragraph and textbox prediction, and 0.35 for image and table prediction, optimizing sensitivity and specificity.
  \item \textbf{imgsz:} Uniformly set at 640 for consistent image processing.
  \item \textbf{stream:} Activated streaming mode for efficient prediction operations.
\end{itemize}

This multifaceted predictive approach underlined our commitment to capturing nuanced document elements, contributing to an advanced and precise document analysis framework.

\subsubsection{\textbf{Post Processing}}
To further enhance the prediction of the models we applied the various post processing on the predictions of the two models.

\textbf{Table mask refinement using convex-hull filling}

Predicted table masks frequently exhibited gaps, at intersections with other classes as most tables have other elements in them.  To address this, a technique involving the use of convex hull filling \cite{convex_hull} was employed. This method effectively mitigated the gaps in table masks, enhancing the overall accuracy and completeness of document element segmentation.

\textbf{Image Mask Prediction}

A second prediction phase segments image masks using a specialized model when the general model detects an image instance in a document. This addresses gaps often found in general model's image masks, especially at intersections with other classes. Firstly, the documents in which the general model detect images are given to the image model. The image model then gives refined prediction of image instances on documents.
Image instance detection by the general model is used to avoid redundant inference through the specialized model, optimizing efficiency.
The confidence threshold of this model was 0.35.

\subsubsection{\textbf{Handling CUDA out of memory exceptions}}

During prediction, certain images led to CUDA out-of-memory (OOM) exceptions, disrupting normal processing and giving empty predictions \cite{cuda}. To address this, a systematic approach was adopted. When a CUDA OOM exception occurred, the "retina\_masks" parameter was dynamically disabled during prediction. This strategic decision prevented further disruptions by alleviating the additional memory burden imposed by retina masks. Images triggering CUDA OOM errors were processed with "retina\_masks" set to "False." This resulted in getting high resolution masks in as many images as possible.

\section{Results \& Discussions,}
Our approach resulted in an iterative improvement in the public DICE\cite{dice} score on Kaggle. In the table:I and table:II we show the improvements of our approach. Furthermore, we present an example of our post processing in figure 2.
\begin{table}[h]
\centering
\begin{tabularx}{9cm}{|X|X|}
\hline
\textbf{Solution} & \textbf{Public DICE Score on Kaggle} \\
\hline
General Model & 0.87291 \\
\hline
General Model + Table Filling & 0.87965 \\
\hline
\end{tabularx}
\caption{Improvements due to table filling on finetuned 18 epochs general model}
\label{tab:model_comparison}
\end{table}

\begin{table}[h]
\centering
\begin{tabularx}{9cm}{|X|X|}
\hline
\textbf{Solution} & \textbf{Public DICE Score on Kaggle} \\
\hline
General Model + Table fill & 0.88907 \\
\hline
General Model + Table fill + Image Model & 0.89024 \\
\hline
General Model + Table fill + Image Model + OOM Handled & 0.89198 \\
\hline
General Model + Table fill + Image Model + OOM Handled + Optimal Configuration & 0.89537 \\
\hline
\end{tabularx}
\caption{Solution Development Performance Comparison on finetuned 50 epoch general model}
\label{tab:model_comparison}
\end{table}

\begin{figure}[!htb]

\centering
\includegraphics[width=.3\textwidth,height=2.5cm,keepaspectratio]{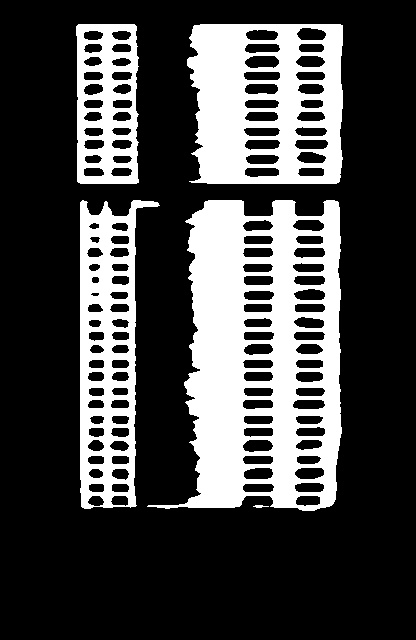}\hfill
\includegraphics[width=.3\textwidth,height=2.5cm,keepaspectratio]{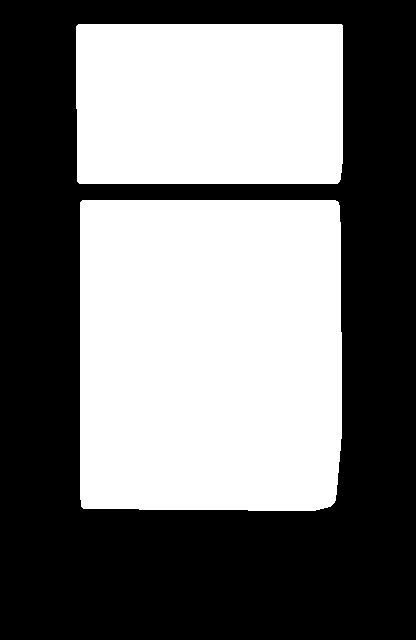}
\includegraphics[width=.3\textwidth,height=2.5cm,keepaspectratio]{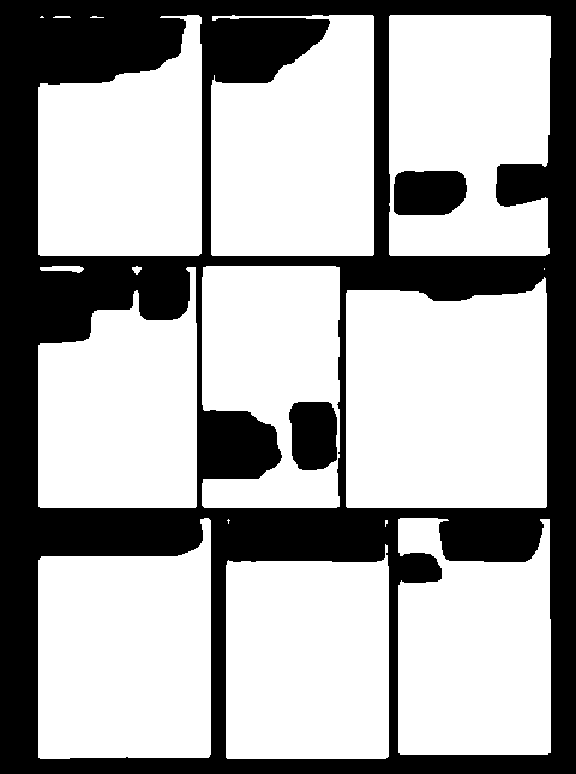}\hfill
\includegraphics[width=.3\textwidth,height=2.5cm,keepaspectratio]{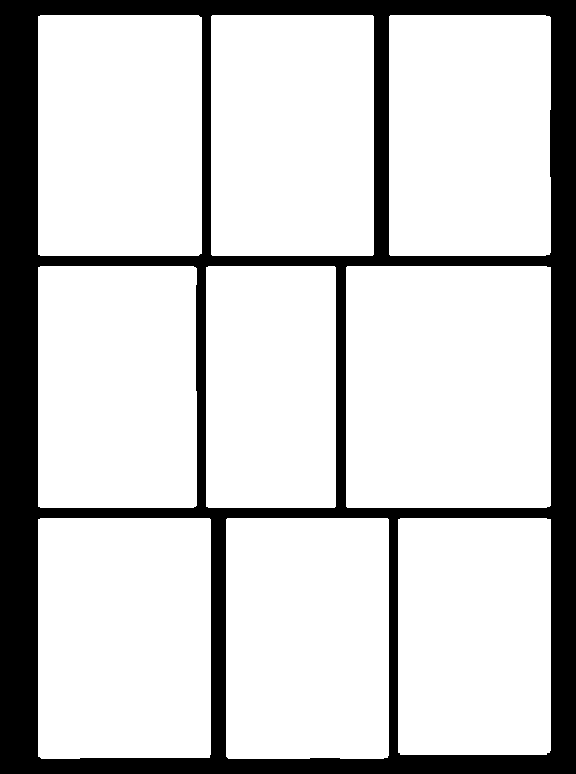}
\caption{Improvement on Tables and Images}
\label{fig:figure3}

\end{figure}

Our proposed approach significantly increased the score. Notably, in our validation testing we saw the DICE scores of images go from \textbf{0.93} to \textbf{0.95 }. 

\section{Conclusion}
In this paper, we presented an ensemble model and post processing based approach for Bengali DLA on the BaDLAD dataset. Our ensemble model consists of using two models(a general and an image model) and mask-filling post-processing using Convex Hull Filling algorithm to accurately produce segment masks.  We showed that our ensemble model outperforms each individual architecture by a significant margin.
Our results demonstrate the effectiveness of the ensemble and post-processing approach for Bengali DLA. Its robustness and efficiency make it usable for practical real time applications. Through a series of iterative improvements and thorough testing, we have harnessed the power of YOLOv8 and other sophisticated techniques to enhance the accuracy and completeness of our segmentation results. 
We believe that our work will make a significant contribution to the field of Bengali DLA. The ensemble model presented in this paper can be used to improve the performance of a variety of document processing tasks, such as OCR and document understanding.


\begin{thebibliography}{00}
\bibitem{dlm_survey_paper} 
C. Doss, S. P. Mohanty, B. M. Mehmood, and A. K. Sangaiah. "A survey on document layout analysis techniques." Artificial Intelligence Review 47.2 (2017): 379-418.
\bibitem{complexity}
Alam, S., Reasat, T., Sushmit, A. S., Siddique, S. M., Rahman, F., Hasan, M., \& Humayun, A. I. (2021, September). A large multi-target dataset of common bengali handwritten graphemes. In International Conference on Document Analysis and Recognition (pp. 383-398). Cham: Springer International Publishing.
\bibitem{badlad}
Hasan, M. M., Arafat, M. H., Iftekhar, K. M., Amin, M. S. M. R., \& Islam, M. S. (2023). BaDLAD: A Large Multi-Domain Bengali Document Layout Analysis Dataset. arXiv preprint arXiv:2303.05325.
\bibitem{yolo}
Redmon, J. et al., 2016. You Only Look Once: Unified, Real-Time Object Detection. In 2016 IEEE Conference on Computer Vision and Pattern Recognition (CVPR). pp. 779–788.
\bibitem{convex_hull}
Barber, C. B., Dobkin, D. P., \& Huhdanpaa, H. (1996). The quickhull algorithm for convex hulls. ACM Transactions on Mathematical Software (TOMS), 22(4), 469-483.
\bibitem{mixup}
Gazda, M., Bugata, P., Gazda, J., Hubacek, D., Hresko, D. J., \& Drotar, P. (2021). Mixup augmentation for kidney and kidney tumor segmentation. In International Challenge on Kidney and Kidney Tumor Segmentation (pp. 90-97). Cham: Springer International Publishing.
\bibitem{copypaste}
Ghiasi, G., Cui, Y., Srinivas, A., Qian, R., Lin, T. Y., Cubuk, E. D., ... \& Zoph, B. (2021). Simple copy-paste is a strong data augmentation method for instance segmentation. In Proceedings of the IEEE/CVF conference on computer vision and pattern recognition (pp. 2918-2928).
\bibitem{mosaic}
Hao, W., \& Zhili, S. (2020, November). Improved mosaic: Algorithms for more complex images. In Journal of Physics: Conference Series (Vol. 1684, No. 1, p. 012094). IOP Publishing.
\bibitem{perspective}
Zhou, W., Zyner, A., Worrall, S., \& Nebot, E. (2019). Adapting semantic segmentation models for changes in illumination and camera perspective. IEEE Robotics and Automation Letters, 4(2), 461-468.
\bibitem{dice}
Setiawan, A. W. (2020, November). Image segmentation metrics in skin lesion: accuracy, sensitivity, specificity, dice coefficient, Jaccard index, and Matthews correlation coefficient. In 2020 International Conference on Computer Engineering, Network, and Intelligent Multimedia (CENIM) (pp. 97-102). IEEE.
\bibitem{map50_score}
Z. Wang, A. C. Berg, T. L. Berg, and J. R. Smith. "Learning to segment object candidates." European Conference on Computer Vision. Springer, Cham, 2013. 179-193.
\bibitem{retina_mask}
F. Ren, M. Sun, B. Dong, and L. Zhang. "RetinaMask: High-Resolution Instance Segmentation with FPN and RetinaNet." arXiv preprint arXiv:1901.03670 (2019).
\bibitem{cuda}
NVIDIA. "CUDA." NVIDIA Developer, 2023. https://developer.nvidia.com/cuda-toolkit.


\end{thebibliography}
\end{document}